\documentclass[letterpaper, 10 pt, conference]{ieeeconf}  

\IEEEoverridecommandlockouts                              

\usepackage{multicol}
\usepackage{graphicx}
\usepackage{caption}      
\usepackage{subcaption}   
\usepackage{wrapfig}
\usepackage{csvsimple}
\usepackage{float}
\usepackage{pgfplotstable,filecontents}
\usepackage{amsmath}
\usepackage{algorithm}
\usepackage[noend]{algpseudocode}

\providecommand{\tabularnewline}{\\}
\usepackage[T1]{fontenc}
\usepackage[latin9]{inputenc}
\usepackage{array}
\usepackage{multirow}

\title{\LARGE \bf
Shape Completion Enabled Robotic Grasping*
}

\author{Jacob Varley, Chad DeChant, Adam Richardson, Joaqu\'in Ruales, and Peter Allen
\thanks{Authors are with Columbia University Robotics Group, 
        Columbia University, New York, NY 10027, USA
        {\tt\small jvarley@cs.columbia.edu, dechant@cs.columbia.edu, ajr2190@columbia.edu, asn2129@columbia.edu, jar2262@columbia.edu, allen@cs.columbia.edu}}
}

\author{Jacob Varley, Chad DeChant, Adam Richardson, Joaqu\'in Ruales, and Peter Allen
\thanks{*This work is supported by NSF Grant IIS-1208153. Thanks to the NVIDIA Corporation for the Titan X GPU grant.}
\thanks{Authors are with Columbia University Robotics Group, 
        Columbia University, New York, NY 10027, USA
        {\tt\small jvarley@cs.columbia.edu, dechant@cs.columbia.edu, ajr2190@columbia.edu, jar2262@columbia.edu, allen@cs.columbia.edu}}%
}

\pgfplotsset{compat=1.13} 

\begin{document}

\maketitle
\thispagestyle{empty}
\pagestyle{empty}

\begin{abstract}

This work provides an architecture to enable robotic grasp planning via shape completion. Shape completion is accomplished through the use of a 3D convolutional neural network (CNN). The network is trained on our own new open source dataset of over 440,000 3D exemplars captured from varying viewpoints. At runtime, a 2.5D pointcloud captured from a single point of view is fed into the CNN, which fills in the occluded regions of the scene, allowing grasps to be planned and executed on the completed object. Runtime shape completion is very rapid because most of the computational costs of shape completion are borne during offline training. We explore how the quality of completions vary based on several factors. These include whether or not the object being completed existed in the training data and how many object models were used to train the network. We also look at the ability of the network to generalize to novel objects allowing the system to complete previously unseen objects at runtime. Finally, experimentation is done both in simulation and on actual robotic hardware to explore the relationship between completion quality and the utility of the completed mesh model for grasping.

\end{abstract}


\section{INTRODUCTION}

Grasp planning based on raw sensory data is difficult due to occlusion and incomplete information regarding scene geometry. This work utilizes 3D convolutional neural networks (CNNs)\cite{lecun} to enable stable robotic grasp planning via shape completion. The 3D CNN is trained to do shape completion from a single pointcloud of a target object, essentially filling in the occluded portions of objects. This ability to infer occluded geometries can be applied to a multitude of robotic tasks. It can assist with path planning for both arm motion and robot navigation where an accurate understanding of whether occluded scene regions are occupied or not results in better trajectories. It also allows a traditional grasp planner to generate stable grasps via the completed shape.

The proposed framework consists of two stages: a training stage and a runtime stage. During the training stage, the CNN is shown occupancy grids created from thousands of synthetically rendered depth images of different mesh models. Each of these occupancy grids is captured from a single point of view, and occluded portions of the volume are marked as empty. For each training example, the ground truth occupancy grid (the occupancy grid for the entire 3D volume) is also generated for the given mesh. From these pairs of occupancy grids the CNN learns to quickly complete mesh models at runtime using only information from a single point of view. Several example completions are shown in 
\begin{figure}[H]
\centering
\includegraphics[width=.45\textwidth] {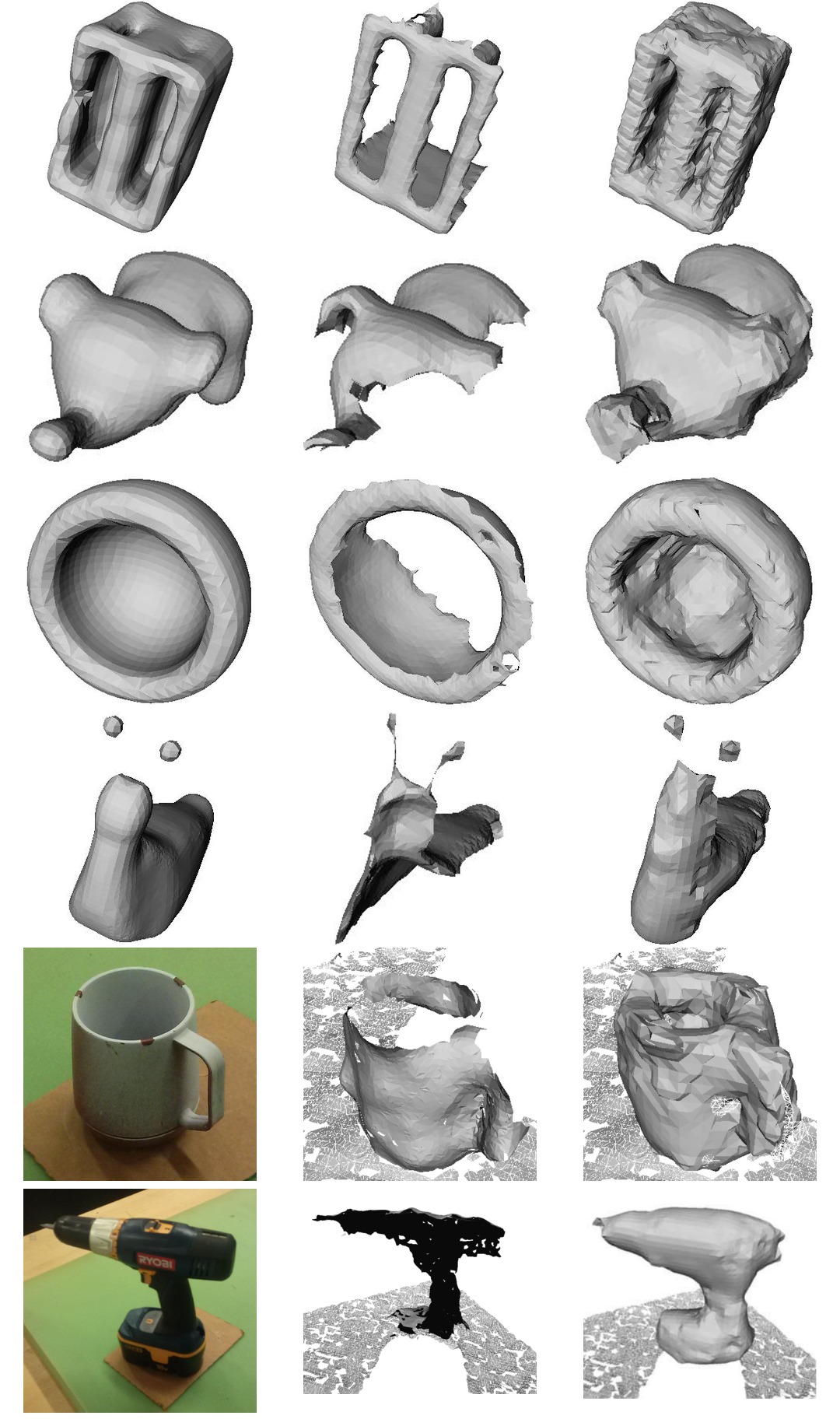}
\caption{Ground Truth, Partials, and Completions (L to R). The top four are completions of synthetic depth images of Grasp Dataset holdout models. The mug and drill show completions of Kinect-captured depth images of physical objects.}
\label{fig:completion_montage}
\end{figure}
\noindent Fig. \ref{fig:completion_montage}. This setup is beneficial for robotics applications as the majority of the computation time takes place during offline training, so that at runtime an object's partial-view pointcloud can be run through the CNN and completed in under a tenth of a second on average and then quickly meshed.

During the runtime stage, a pointcloud is captured using a depth sensor. A segmentation algorithm is run, and regions of the pointcloud corresponding to graspable objects are extracted from the scene. Occupancy grids of these regions are created, where all occluded regions are labeled as empty. These maps are passed separately through the trained CNN. The outputs from the CNN are occupancy grids, where the CNN has labeled all the occluded regions of the input as either occupied or empty for each object. These new occupancy grids are either run through a fast marching cubes algorithm\cite{lorensen1987marching}, or further post-processed if they are to be grasped. Whether the object is completed or completed and post-processed results in either 1) fast completions suitable for path planning and scene understanding or 2) detailed meshes suitable for grasp planning, where the higher resolution visible regions are incorporated into the reconstruction. This framework is extensible to crowded scenes with multiple objects as each object is completed individually. It is also applicable to different domains because it can learn to reproduce objects from whatever dataset it is trained on, and further shows the ability to generalize to unseen views of objects or even entirely novel objects. This applicability to multiple domains is complemented by the thousands of 3D models available from datasets such as ShapeNet\cite{chang2015shapenet} and the rapidly increasing power of GPU processors. 

The contributions of this work include: 1) A novel CNN architecture for shape completion; 2) A fast mesh completion method, resulting in meshes able to quickly fill the planning scene; 3) A second CUDA enabled completion method that creates detailed meshes suitable for grasp planning by integrating fine details from the observed pointcloud; 4) A large open-source dataset of over 440,000 $40^3$ voxel grid pairs used for training. This dataset and the related code are freely available at {\bf http://shapecompletiongrasping.cs.columbia.edu}. In addition, the website makes it easy to browse and explore the thousands of completions and grasps related to this work; 5) Results from both simulated and live experiments comparing our method to other approaches and demonstrating its improved performance in grasping tasks.




\section{Related Work}

General shape completion to enable robotic grasping has been studied in robotics. Typical approaches \cite{bohg2011mind}\cite{quispe2015exploiting}\cite{schiebener2016heuristic} use symmetry and extrusion heuristics for shape completion, and they are reasonable for objects well represented by geometric primitives. Our approach differs from these methods in that it learns to complete arbitrary objects based upon a large set of training exemplars, rather than requiring the objects to conform to heuristics.

A common alternative to general shape completion in the robotics community is object recognition and 3D pose detection \cite{rennie2016dataset}\cite{papazov2010efficient}\cite{hinterstoisser2011multimodal}. In these approaches objects are recognized from a database of objects and the pose is then estimated. These techniques fill a different use case: the number of encountered objects is small, and known ahead of time often in terms of both texture and geometry. Our approach differs in that it extends to novel objects. 

The computer vision and graphics communities have become increasingly interested in the problem of shape completion. Some examples include \cite{wu20143d}\cite{wu20153d}, which use a deep belief network and Gibbs sampling for 3D shape reconstruction, and \cite{firman2016structured}, which uses Random Forests. In addition, work by \cite{rock2015completing} uses an exemplar based approach for the same task. Others \cite{tulsiani2016learning}\cite{choy20163d} have developed algorithms to learn 3D occupancy grids directly from 2D images. Li et. al \cite{li2015Database} uses a database of pre-existing models to do completion.
\begin{wrapfigure}{r}{0.25\textwidth}
    \includegraphics [width=0.25\textwidth] {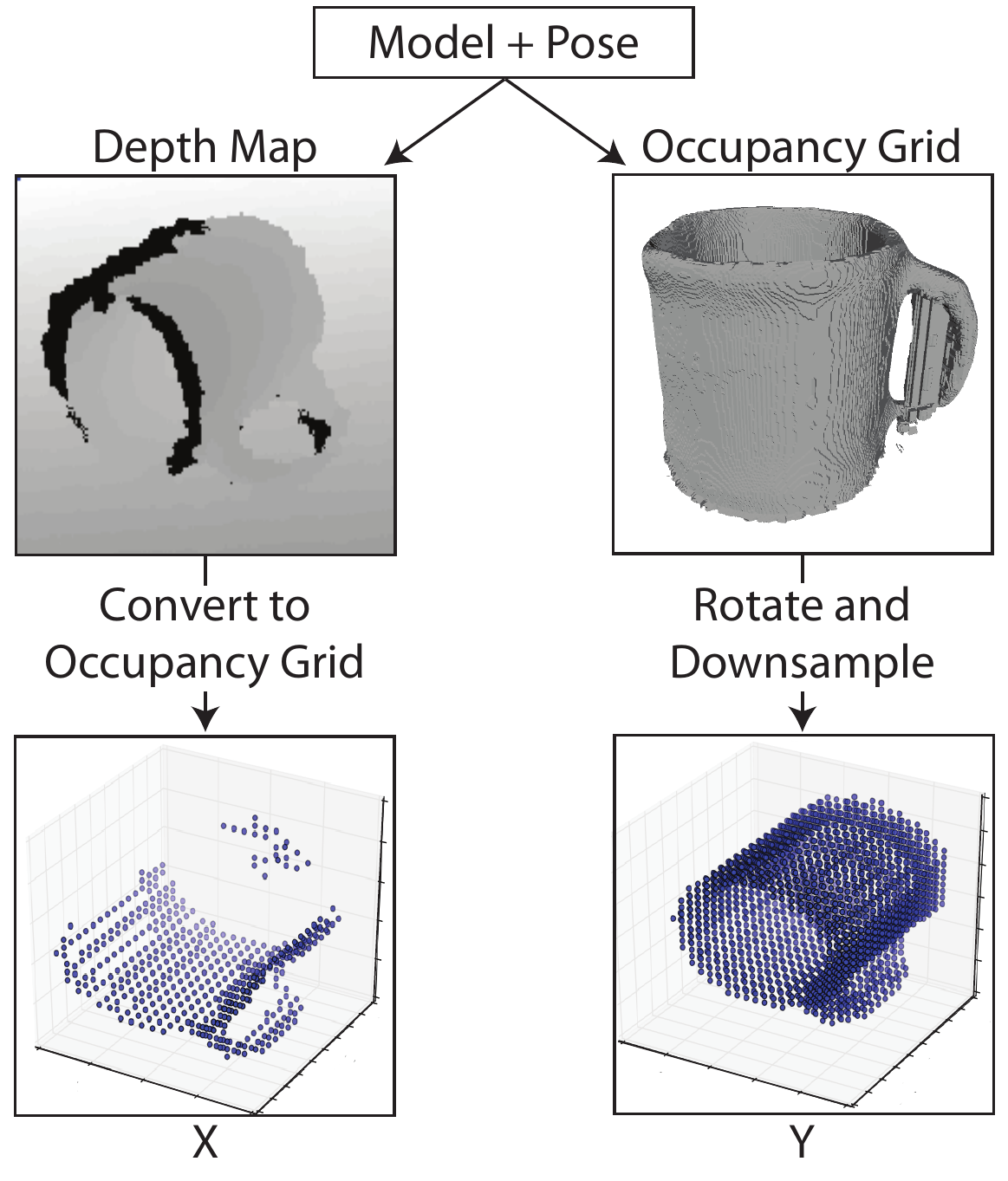}
  \caption{Training Data; In X, the input to the CNN, the occupancy grid marks visible portions of the model. Y, the expected output, has all voxels occupied by the model marked.}
  \label{fig:data_generation}
\end{wrapfigure}

It is difficult to apply many of these works directly to robotic manipulation as no large dataset of renderings of handheld objects needed for robotic manipulation tasks existed until now. Also, \cite{tulsiani2016learning}\cite{choy20163d} use pure RGB rather than the RGBD images prevalent in robotics, making the problem more difficult as the completed shape must somehow be positioned in the scene and the process does not utilize available depth information. Most create results with resolutions too low for use with current grasp planners which require meshes. Our work creates a new dataset specifically designed for completing objects useful for manipulation tasks using the 2.5-D range sensors prevalent in robotics, and provides a technique to integrate the high resolution observed view of the object with our relatively high resolution CNN output, creating a completion suitable for grasp planning. 

Our work differs from \cite{mahler2016dex} and our own related work \cite{goldfeder2009columbia}, both of which require a complete mesh to query a model database and retrieve grasps used on similar objects. These approaches could be used in tandem with our framework where the completed model would act as the query mesh. While grasps can be planned using partial meshes where the object is not completed (see \cite{varley2015generating}), they still have their limitations and issues. Shape completion can be used to alleviate this problem.

While many mesh model datasets exist such as \cite{chang2015shapenet}, \cite{wu20143d}, and \cite{xiang_wacv14}, this framework makes heavy use of the YCB\cite{calli2015ycb} and Grasp Database\cite{kappler2015leveraging} mesh model datasets. We chose these two datasets as many robotics labs all over the world have physical copies of the YCB objects, and the Grasp Database contains objects specifically geared towards robotic manipulation. We augmented 18 of the YCB objects whose provided meshes were of high quality with models from the Grasp Database which contains 590 mesh models.

\begin{figure*}
\vspace{2mm}
\centering
\includegraphics  [width=0.95\textwidth] {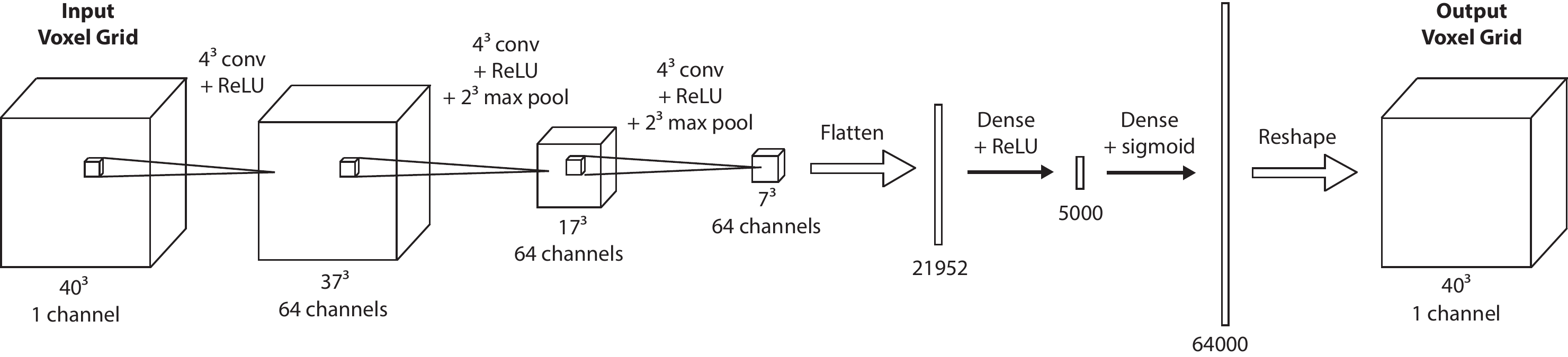}
\caption{CNN Architecture. The CNN has three convolutional and two dense layers. The final layer has $64000$ nodes, and reshapes to form the resulting $40^3$ occupancy grid. The numbers on the bottom edges show the input sizes for each layer. All layers use ReLU activations except for the last dense layer, which uses a sigmoid.}
\label{fig:model_architecture}
\end{figure*}

\begin{figure*}[t]
\centering
	\begin{subfigure}[t]{2.2in}
		\centering
		\includegraphics[width=2.2in]{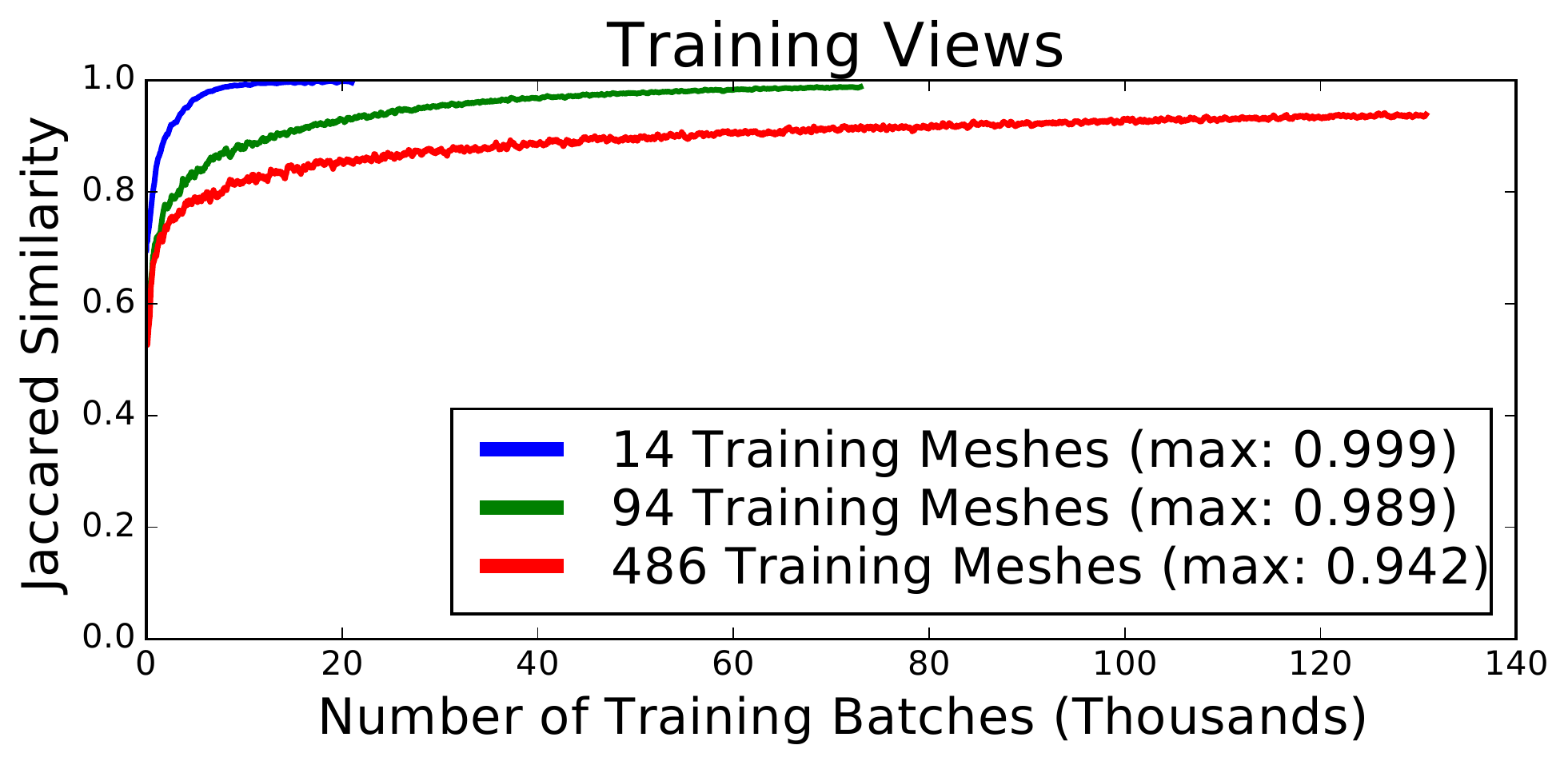}
		\caption{}
	\end{subfigure}
	\begin{subfigure}[t]{2.2in}
		\centering
		\includegraphics[width=2.2in]{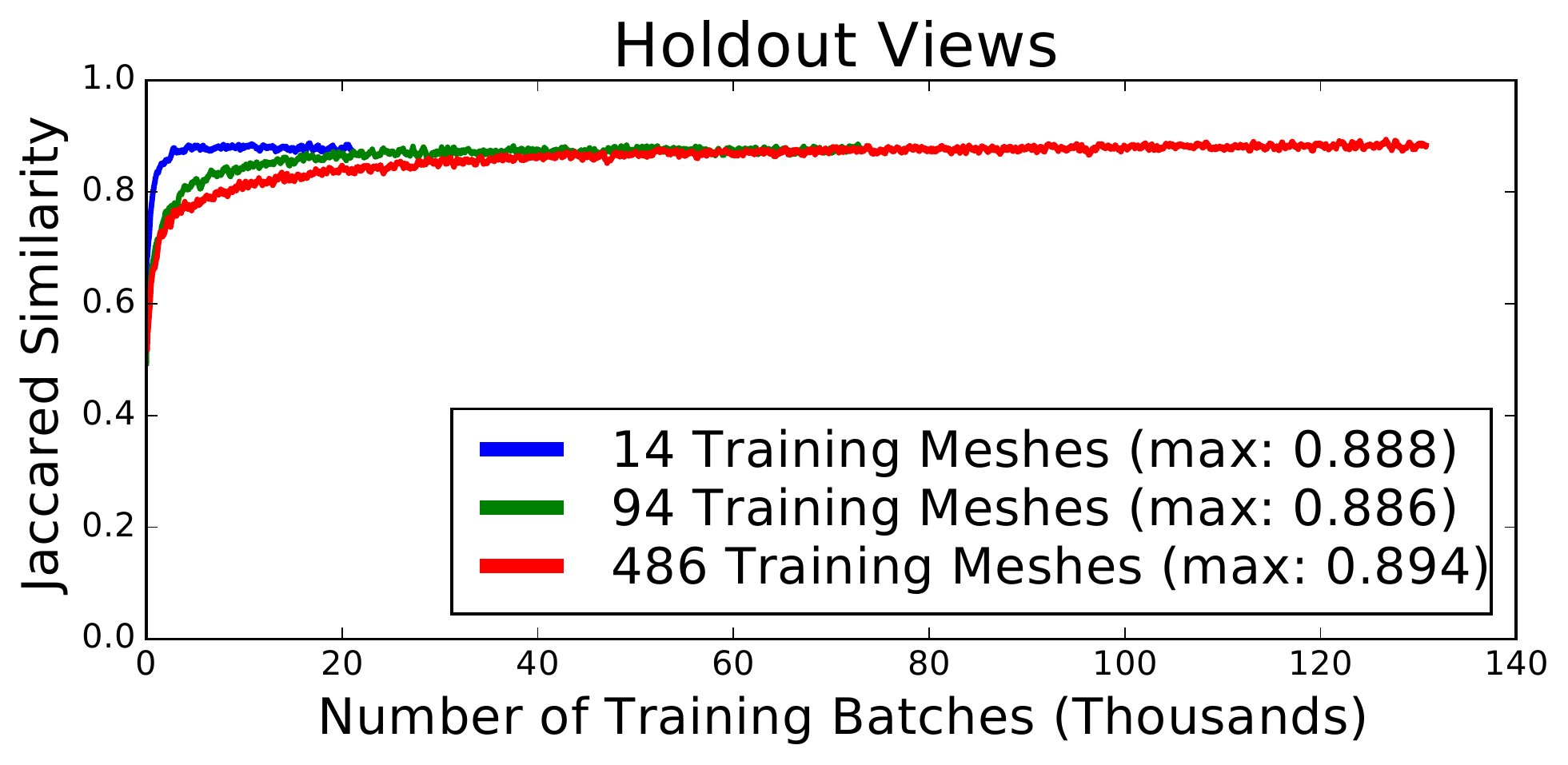}
		\caption{}
	\end{subfigure}
	\begin{subfigure}[t]{2.2in}
		\centering
		\includegraphics[width=2.2in]{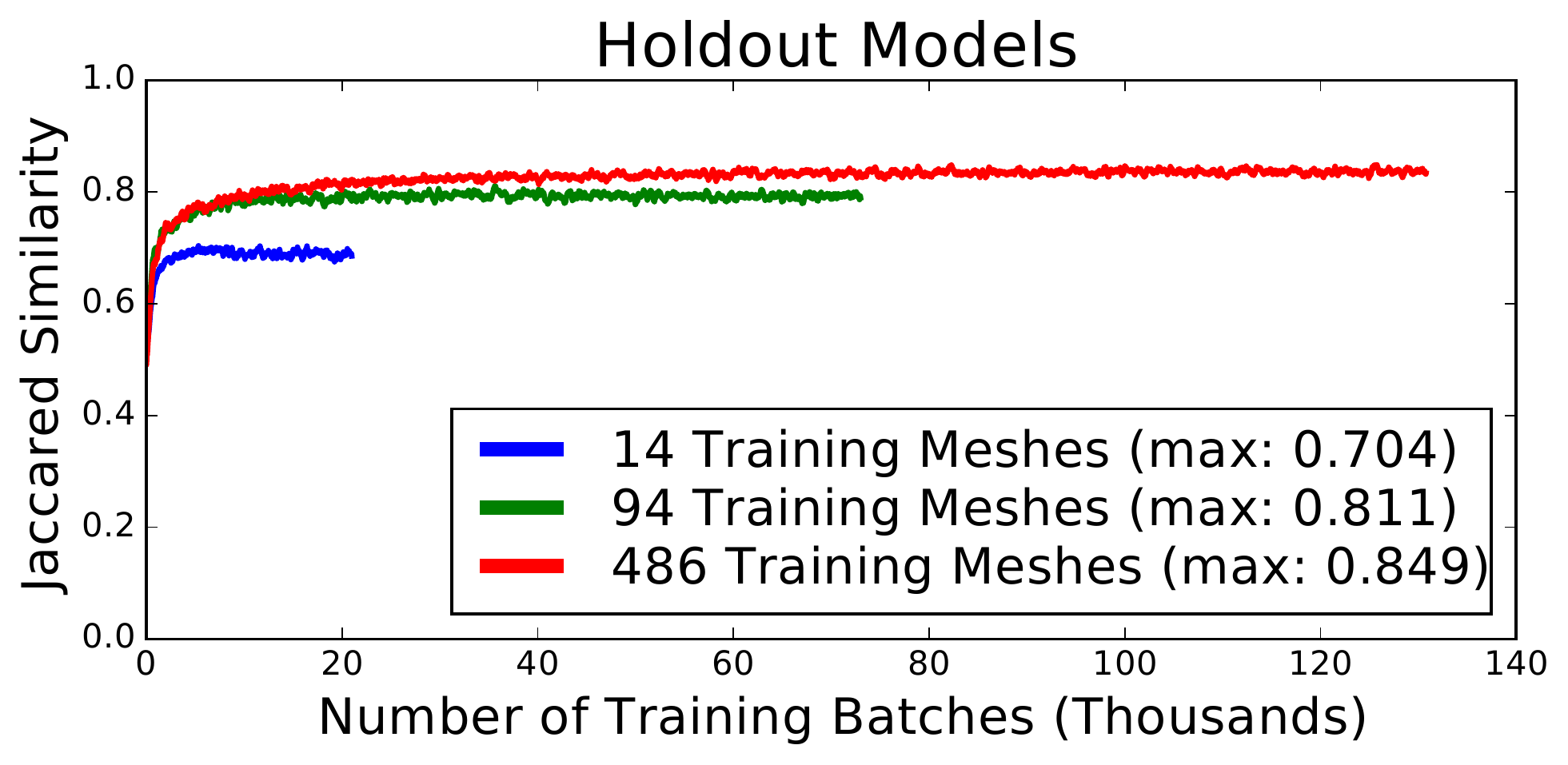}
		\caption{}
	\end{subfigure}
\caption{Jaccard similarity for three CNNs, one (shown in blue) trained with 14 mesh models, the second (green) trained with 94 mesh models, and the third (red) trained with 486 mesh models. For each plot, while training, the CNNs were evaluated on inputs they were being trained on (Training Views, plot a), novel inputs from meshes they were trained on (Holdout Views, plot b) and novel inputs from meshes they have never seen before (Holdout Models, plot c).} 
\label{fig:jaccard_plots} 
\end{figure*} 


\section{Training}

\subsection{Data Generation}

In order to train a network to reconstruct a diverse range of objects, meshes were collected from the YCB and Grasp Database. The models were run through binvox\cite{min2004binvox} in order to generate $256^3$ occupancy grids. In these occupancy grids, both the surface and interior of the meshes are marked as occupied. In addition, all the meshes were placed in Gazebo\cite{koenig2004design}, and 726 depth images were generated for each object subject to different rotations uniformly sampled (in roll-pitch-yaw space, 11*6*11) around the mesh. The depth images are used to create occupancy grids for the portions of the mesh visible to the simulated camera, and then all the occupancy grids generated by binvox are transformed to correctly overlay the depth image occupancy grids. Both sets of occupancy grids are then down-sampled to $40^3$ to create a large number of training examples. The input set (X) contains occupancy grids that are filled only with the regions of the object visible to the camera, and the output set (Y) contains the ground truth occupancy grids for the space occupied by the entire model. An illustration of this process is shown in Fig. \ref{fig:data_generation}.

\subsection{Model Architecture and Training}

The architecture of the CNN is shown in Fig. \ref{fig:model_architecture}. The model was implemented using Keras\cite{Keras2015}, a Theano\cite{bergstra2010theano}\cite{bastien2012theano} based deep learning library. Each layer used rectified linear units as nonlinearities except the final fully connected (output) layer which used a sigmoid activation to restrict the output to the range $[0,1]$. We used the cross-entropy error $E(y,y^\prime)$ as the cost function with target $y$ and output $y^\prime$:
$$E(y,y^\prime)=-\left( y \log(y^\prime) + (1 - y) \log(1 - y^\prime) \right) $$

\noindent This cost function encourages each output to be close to either 0 for unoccupied target voxels or 1 for occupied. The optimization algorithm Adam\cite{kingma2014}, which computes adaptive learning rates for each network parameter, was used with default hyperparameters ($\beta_1$$=$$0.9$, $\beta_2$$=$$0.999$, $\epsilon$$=$$10^{-8}$) except for the learning rate, which was set to 0.0001. Weights were initialized following the recommendations of \cite{he2015} for rectified linear units and \cite{glorot2010} for the logistic activation layer. The model was trained with a batch size of 32. Each of the 32 examples in a batch was randomly sampled from the full training set with replacement.

We used the Jaccard similarity to evaluate the similarity between a generated voxel occupancy grid and the ground truth. The Jaccard similarity between sets A and B is given by:
$$J(A,B)=\frac{|A\cap B|}{|A\cup B|}$$
\noindent The Jaccard similarity has a minimum value of 0, where A and B have no intersection and a maximum value of 1 where A and B are identical. During training, this similarity measure is computed for input meshes that were in the training data ({\bf  Training Views}), meshes from objects within the training data but from novel views ({\bf  Holdout Views}), and for meshes of objects not in the training data ({\bf Holdout Models}). The CNNs were trained with an NVIDIA Titan X GPU.

\begin{figure*}[t]
\vspace{2.5mm}
\centering
	\begin{subfigure}[t]{1.65in}
		\centering
		\includegraphics[width=1.65in, height=1.5in]{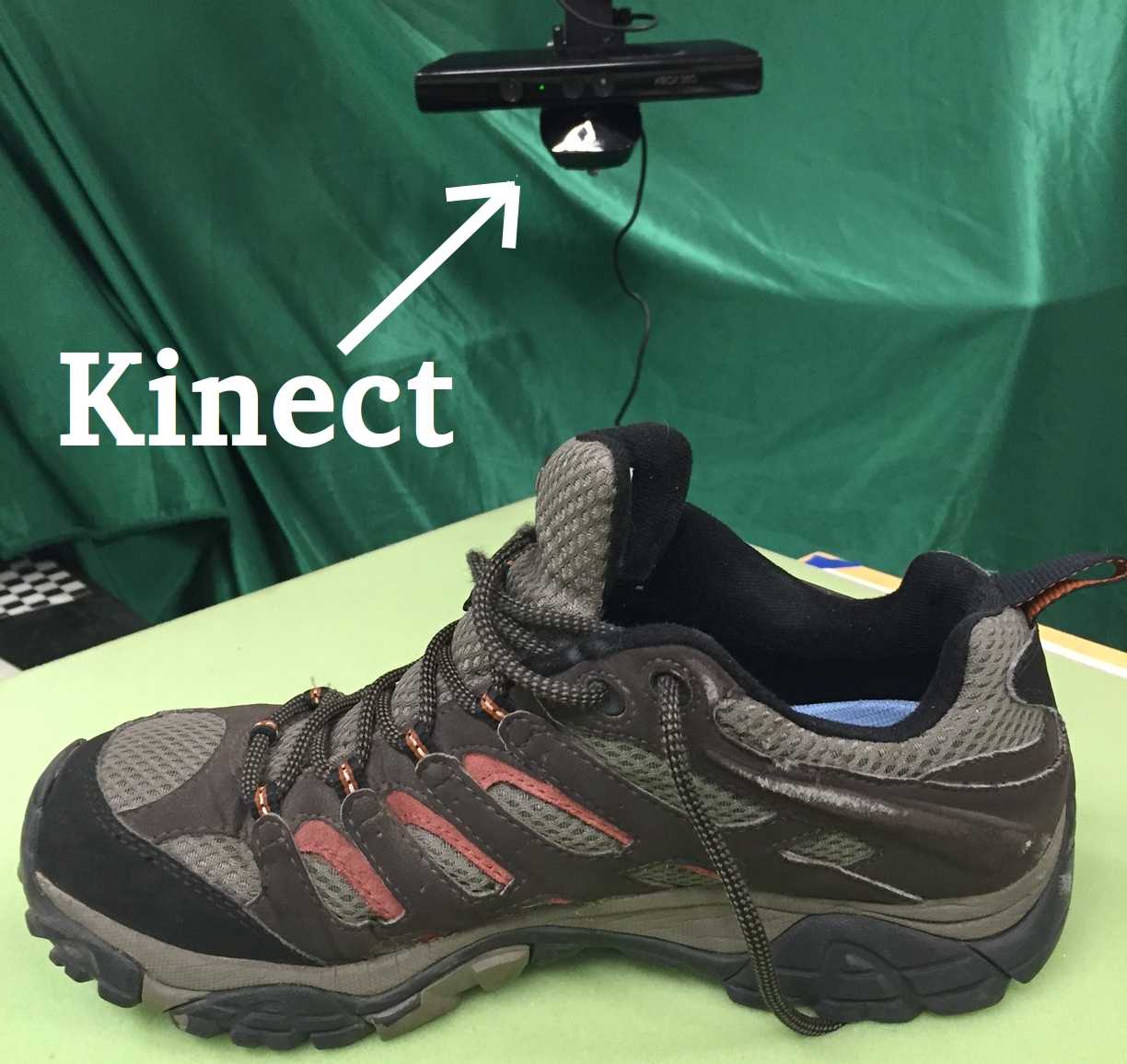}
		\caption{Image of Occluded Side}\label{fig:ball_rgb}	
	\end{subfigure}
	\begin{subfigure}[t]{1.65in}
		\centering
		\includegraphics[width=1.65in, height=1.5in]{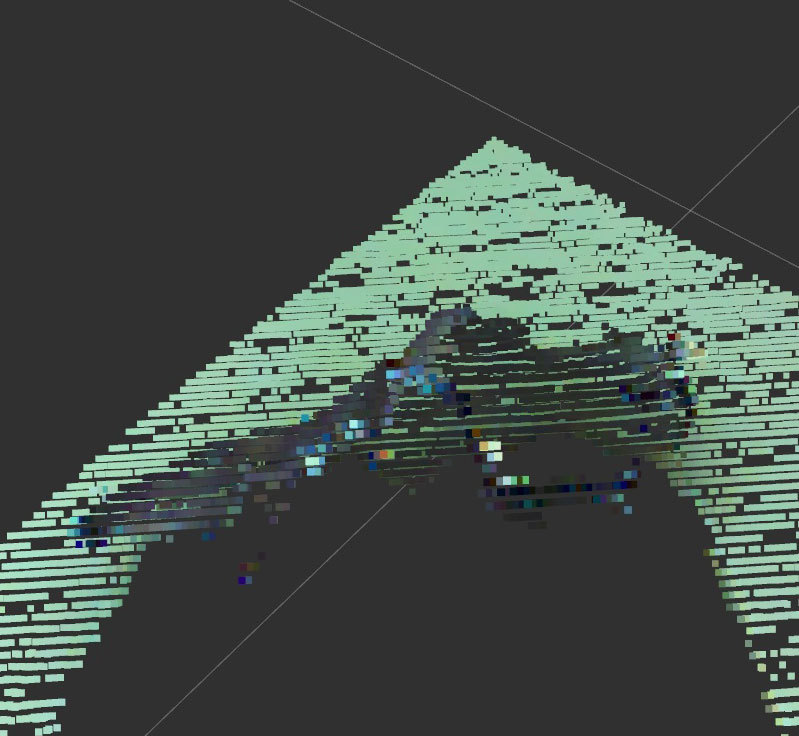}
		\caption{Point Cloud}\label{fig:ball_pc_crop}
	\end{subfigure}
	\begin{subfigure}[t]{1.65in}
		\centering
		\includegraphics[width=1.65in, height=1.5in]{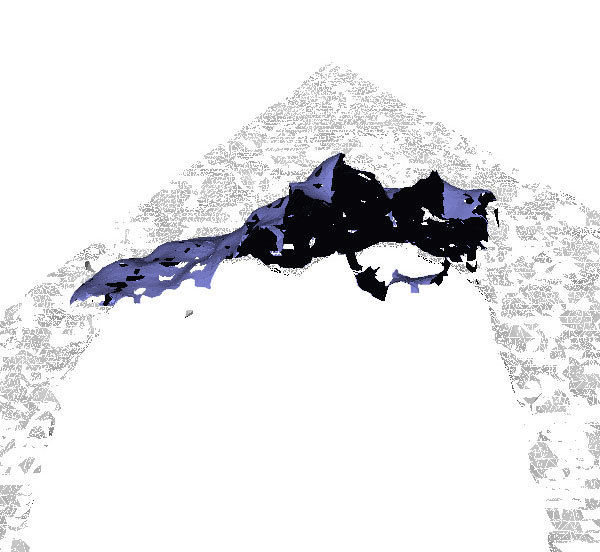}
		\caption{Segmented and Meshed}\label{fig:ball_size_view_segmented}	
	\end{subfigure}
	\begin{subfigure}[t]{1.65in}
		\centering
		\includegraphics[width=1.65in, height=1.5in]{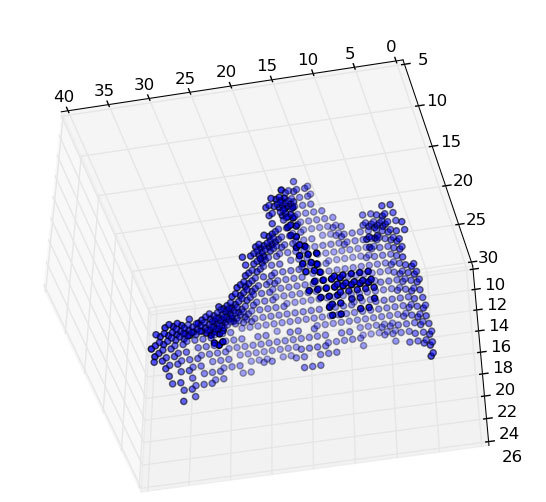}
		\caption{CNN Input}\label{fig:ball_cnn_input}
	\end{subfigure}
		\begin{subfigure}[t]{1.65in}
		\centering
		\includegraphics[width=1.65in, height=1.5in]{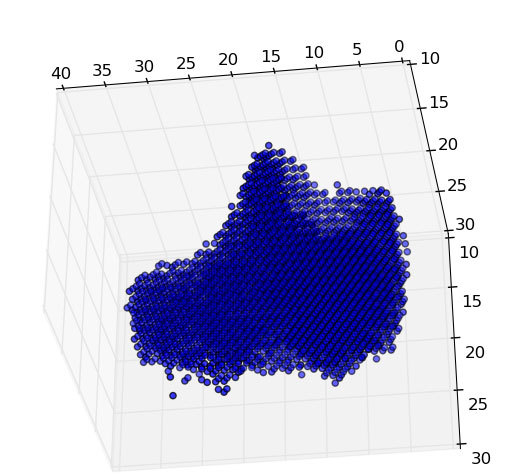}
		\caption{CNN Output}\label{fig:ball_cnn_output}	
	\end{subfigure}
	\begin{subfigure}[t]{1.65in}
		\centering
		\includegraphics[width=1.65in, height=1.5in]{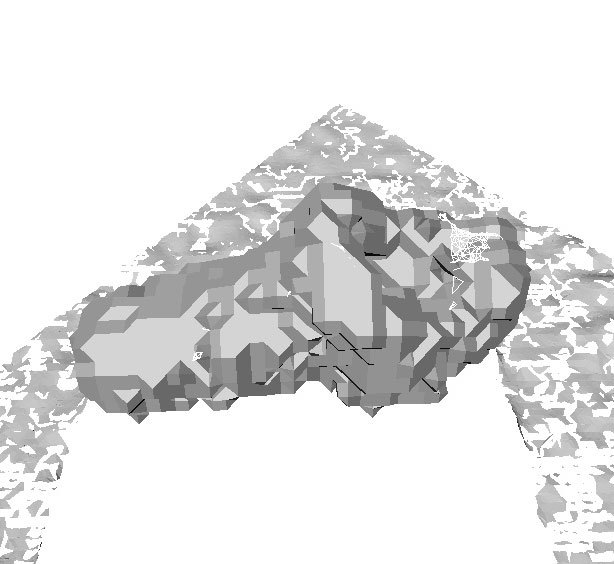}
		\caption{Fast Mesh}\label{fig:ball_mcubes}
	\end{subfigure}
		\begin{subfigure}[t]{1.65in}
		\centering
		\includegraphics[width=1.65in, height=1.5in]{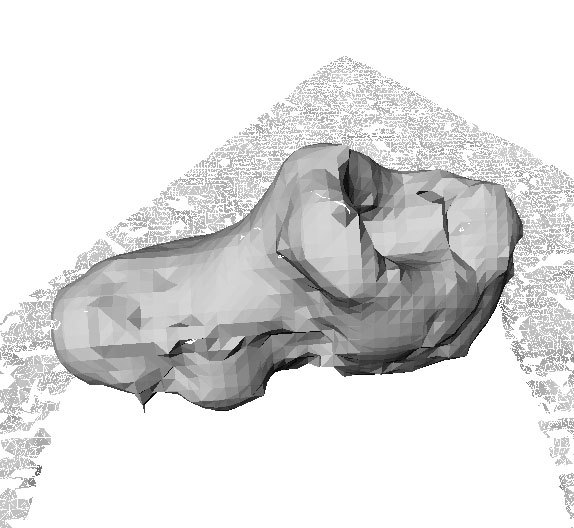}
		\caption{Detailed Mesh}\label{fig:ball_smoothed}
	\end{subfigure}
	\begin{subfigure}[t]{1.65in}
		\centering
		\includegraphics[width=1.65in, height=1.5in]{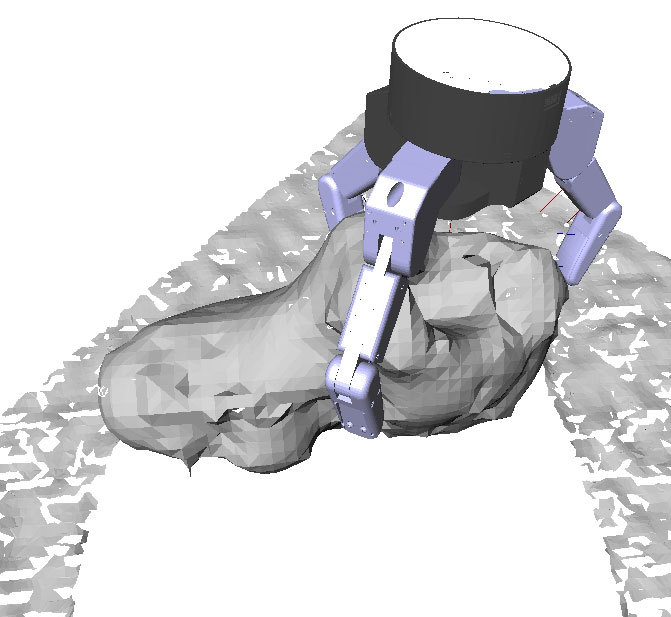}
		\caption{Grasp Planning}\label{fig:ball_grasp}
	\end{subfigure}

\caption{Stages to the Runtime Pipeline. These images are not shown from the angle in which the data was captured in order to visualize the occluded regions. (a): An object to be grasped is placed in the scene. (b): A pointcloud is captured. (c): The pointcloud is segmented and meshed. (d): A partial mesh is selected by the user and then voxelized and passed into the 3D shape completion CNN. (e): The output of the CNN. (f): The resulting occupancy grid can be run through a marching cubes algorithm to obtain a mesh quickly. (g): Or, for better results, the output of the CNN can be combined with the observed pointcloud and preprocessed for smoothness before meshing. (h): Grasps are planned on the smoothed completed mesh. Note: this is a novel object not seen by the CNN during training.} 
\label{fig:runtime_pipeline} 
\end{figure*}

\subsection{Training Results}
Fig. \ref{fig:jaccard_plots} shows how the Jaccard similarity measures vary as the networks' training progresses. In order to explore how the quality of the reconstruction changes as the number of models in the training set is adjusted, we trained three networks with identical architectures using variable numbers of mesh models. One was trained with partial views from 14 YCB models, another with 94 mesh models (14 YCB + 80 Grasp Database), and the third with 486 mesh models (14 YCB models + 472 Grasp Database). Each network was allowed to train until learning plateaued; for the CNN trained on 486 objects, this took over a week. The remaining 4 YCB and 118 Grasp Dataset models were kept as a holdout set. Results are shown in Fig. \ref{fig:jaccard_plots}. We note that the networks trained with fewer models perform better shape completion when they are tested on views of objects they have seen during training than networks trained on a larger number of models. This suggests that the network is able to completely learn the training data for the smaller number of models but struggles to do so when trained on larger numbers. Conversely, the models trained on a larger number of objects perform better than those trained on a smaller number when asked to complete novel objects. Because, as we have seen, the networks trained on larger numbers of objects are unable to learn all of the models seen in training, they may be forced to learn a more general completion strategy that will work for a wider variety of objects, allowing them to better generalize to objects not seen in training. 

Fig. \ref{fig:jaccard_plots}(a) shows the performance of the three CNNs on training views. In this case, the fewer the mesh models used during training, the better the completion results. Fig. \ref{fig:jaccard_plots}(b) shows how the CNNs performed on novel views of the mesh objects used during training. Here the CNNs all did approximately the same. Fig. \ref{fig:jaccard_plots}(c) shows the completion quality of the CNNs on objects they have not seen before. In this case, as the number of mesh models used during training increases, performance improves as the system has learned to generalize to a wider variety of inputs.


\section{Runtime}

At runtime the pointcloud for the target object is acquired from a 3D sensor, scaled, voxelized and then passed through the CNN. The output of the CNN, a completed voxel grid of the object, goes through a post processing algorithm that returns a mesh model of the completed object. Finally, a grasp can be planned and executed based on the completed mesh model. Fig. \ref{fig:runtime_pipeline} demonstrates the full runtime pipeline on a novel object never seen before.

\noindent\textbf{1) Acquire Target Pointcloud:} First, a pointcloud is captured using a Microsoft Kinect, then segmented using PCL's\cite{Rusu_ICRA2011_PCL} implementation of euclidean cluster extraction. A segment corresponding to the object to be completed is selected either manually or automatically and passed it to the shape completion module. \\
\textbf{2) Complete via CNN:} The selected pointcloud is then used to create an occupancy
grid with resolution $40^3$. This occupancy grid is used as input to the CNN whose
output is an equivalently sized occupancy grid for the completed shape. In order to fit 
the pointcloud to the $40^3$ grid, it is scaled down uniformly so that the bounding box
of the pointcloud fits in a $32^3$ voxel cube, and then centered in the $40^3$ grid such that the center of the bounding box is at point (20, 20, 18) in the voxel grid. Finally all voxels occupied by points from this scaled and transformed pointcloud are marked as such. Placing the pointcloud slightly off-center in the z dimension leaves more space in the back half of the grid for the network to fill. \\
\textbf{3a) Create Fast Mesh:} At this point, if the object being completed is not going to be grasped, then the voxel grid output by the CNN is run through the marching cubes algorithm, and the resulting mesh is added to the planning scene, filling in occluded regions of the scene. \\
\textbf{3b) Create Detailed Mesh:} Alternatively, if this object is going to be grasped, then post-processing occurs. The purpose of this post-processing is to integrate the points from the visible portion of the object with the output of the CNN. This partial view is of much higher density than the $40^3$ grid and captures significantly finer detail for the visible surface. This merge is made difficult by the large disparity in point densities between the captured cloud and $40^3$ CNN output which can lead to holes and discontinuities if the points are naively merged and run through marching cubes. 

\begin{algorithm}
\caption{Shape Completion}\label{alg:reconstruction}
\begin{algorithmic}[1]
\Procedure{Mesh}{cnn\_out, observed\_pc}
\State //cnn\_out: $40^3$ voxel output from CNN
\State //observed\_pc: captured pointcloud of object
\If{FAST} \Return mCubes(cnn\_out) \EndIf
\State d\_ratio~$\gets$~densityRatio(observed\_pc, cnn\_out) \label{line:densityRatio}
\State upsampled\_cnn~$\gets$~upsample(cnn\_out, d\_ratio) \label{line:upsample}
\State vox~$\gets$~merge(upsampled\_cnn, observed\_pc) \label{line:merge}
\State vox\_no\_gap~$\gets$~fillGaps(vox) \label{line:fill_gaps}
\State vox\_weighted~$\gets$~CUDA\_QP(vox\_no\_gap) \label{line:cuda_qp}
\State mesh~$\gets$~mCubes(vox\_weighted) \label{line:mcubes}
\State \Return mesh
\EndProcedure
\end{algorithmic}
\end{algorithm}

Alg. \ref{alg:reconstruction} shows how we integrated the dense partial view with our $40^3$ voxel grid via the following steps. (Alg.\ref{alg:reconstruction}:L\ref{line:densityRatio}) In order to merge with the partial view, the output of the CNN is converted to a point cloud and its density is compared to the density of the partial view point cloud. The densities are computed by randomly sampling $\frac{1}{10}$ of the points and averaging the distances to their nearest neighbors.
(Alg.\ref{alg:reconstruction}:L\ref{line:upsample}) The CNN output is up-sampled by $d\_ratio$ to match the density of the partial view. This is performed by examining each cube of 8 adjacent original low resolution voxels, with the centers of the voxels as the corners. The new voxels are uniformly distributed inside the cube. For each new voxel, the L1 distance to each original voxel is computed and the 8 distances are summed, weighted by 1 if the original voxel is occupied and -1 otherwise. The new voxel is occupied if its weighted sum is nonnegative. This has the effect of creating piecewise linear separating surfaces similar to the marching cubes algorithm and mitigates up-sampling artifacts. (Alg.\ref{alg:reconstruction}:L\ref{line:merge}) The upsampled output from the CNN is then merged with the point cloud of the partial view and the combined cloud is voxelized at the new higher resolution of $(40*d\_ratio)^3$. For most objects $d\_ratio$ tends to be either 2 or 3, depending on the physical size of the object, resulting in a voxel grid of either $80^3$ or $120^3$. (Alg.\ref{alg:reconstruction}:L\ref{line:fill_gaps}) Any gaps in the voxel grid between the upsampled CNN output and the partial view cloud are filled. This is done by finding the first occupied voxel in every z-stack. If the distance to the next occupied voxel is less than $d\_ratio+1$ the intermediate voxels are filled. (Alg.\ref{alg:reconstruction}:L\ref{line:cuda_qp}) The voxel grid is smoothed using our own CUDA implementation of the convex quadratic optimization problem from \cite{lempitsky2010surface}. This optimization re-weights the voxels, minimizing the Laplacian on the boundary of the embedding function $F$, i.e.: 
$$\int \left( \frac{\partial^2F}{\partial x^2}\right)^2 +\left( \frac{\partial^2F}{\partial y^2}\right)^2 + \left( \frac{\partial^2F}{\partial z^2}\right)^2 dV \to \text{min.}$$
subject to the hard constraint:
$$\forall i,j,k \qquad v_{ijk}\cdot f_{ijk} \geq 0$$
The constraint means that for all input voxels $v$ and output weighted voxels $f$ all occupied voxels prior to the optimization stays occupied or on the boundary, and all unoccupied voxels remain unoccupied or on the boundary.  Again, for further details see \cite{lempitsky2010surface}.
(Alg.\ref{alg:reconstruction}:L\ref{line:mcubes}) The weighted voxel grid is then run through  marching cubes. \\
\textbf{4) Grasp completed mesh:} The reconstructed mesh is then loaded into GraspIt!\cite{miller2004graspit} where a grasp planner is run using a Barrett Hand model. The reachability of the planned grasps are checked using MoveIt!\cite{sucan2013moveit}, and the highest quality reachable grasp is then executed.


\section{Experimental Results}

We created a test dataset by randomly sampling 50 training views ({\bf Training Views}), 50 holdout views ({\bf Holdout Views}), and 50 views of holdout models ({\bf Holdout Models}). The Training Views and Holdout Views were sampled from the 14 YCB training objects. The Holdout Models were sampled from holdout YCB and Grasp Dataset objects.  We used three metrics to compare the accuracy of the different completion methods: Jaccard similarity, Hausdorff distance, and geodesic divergence. 

\subsection{General Completion Results}
\label{sec:Completion_results}
We first compared several general completion methods: passing the partial view through marching cubes and then Meshlab's Laplacian smoothing ({\bf  Partial}), mirroring completion\cite{bohg2011mind} ({\bf Mirror}), our method ({\bf  Ours}). Our CNN was trained on the 484 objects from the YCB + Grasp Dataset and the weights come from the point of peak performance on holdout models (the peak of the red line in Fig. \ref{fig:jaccard_plots}.(c)).

\begin{table}[t]
\vspace{2mm}
\centering
\begin{tabular}{|c|c|c|c|c|}
\hline
View Type       &  Partial  &  Mirror  &  Ours  \\ \hline
Training Views     &  0.1182   &  0.2325  &  \textbf{0.7771}  \\ \hline
Holdout Views   &  0.1307   &  0.2393  &  \textbf{0.7486}   \\ \hline
Holdout Models  &  0.0931   &  0.1921  &  \textbf{0.6496} \\ \hline 
\end{tabular}
\caption{Jaccard Similarity Results (Larger is better). This measures the intersection over union of two voxelized meshes as described in Section \ref{sec:Completion_results}.}
\label{tab:Jaccard}
\end{table}

\begin{table}[t]
\centering
\begin{tabular}{|c|c|c|c|c|}
\hline
View Type      & Partial      & Mirror      & Ours   \\ \hline
Training Views    &  11.4 & 7.5  &  \textbf{3.6}  \\ \hline
Holdout Views  &  12.3 & 8.2  &  \textbf{4.0}  \\ \hline
Holdout Models &  13.6 & 10.7 &  \textbf{5.9}\\ \hline 
\end{tabular}
\caption{Hausdorff Distance Results (Smaller is better). This measures the mean distance in millimeters from points on one mesh to another as described in Section \ref{sec:Completion_results}.}
\label{tab:Hausdorff}
\end{table}

\begin{table}[t]
\vspace{2mm}
\centering
\begin{tabular}{|c|c|c|c|c|}
\hline
View Type  & Partial  & Mirror  & Ours  \\ \hline
Training Views    & 0.3770 & 0.2905 & \textbf{0.0867}   \\ \hline
Holdout Views  & 0.4944 & 0.3366 & \textbf{0.0934} \\ \hline
Holdout Models & 0.3407 & 0.2801 & \textbf{0.1412} \\ \hline
\end{tabular}

\caption{Geodesic Divergence Results (Smaller is better). This measures the Jenson-Shannon probabilistic divergence between two meshes as described in Section \ref{sec:Completion_results}.}
\label{tab:Geodesic}
\end{table}

The Jaccard similarity was used to guide training, as shown in Fig. \ref{fig:jaccard_plots}. We also used this metric to compare the final resulting meshes from several completion strategies. The completed meshes were voxelized at $80^3$, and compared with the ground truth mesh. The results are shown in Table \ref{tab:Jaccard}. Our proposed method results in higher similarity to the ground truth meshes than the partial and mirroring approaches for all tested views.

The Hausdorff distance is a one-directional metric computed by sampling points on one mesh and computing the distance of each sample point to its closest point on the other mesh. The mean value of a completion is the average distance from the sample points on the completion to their respective closest points on the ground truth mesh. The symmetric Hausdorff distance was computed by running Meshlab's\cite{cignoni2008meshlab} Hausdorff distance filter in both directions. Table \ref{tab:Hausdorff} shows the mean values of the symmetric Hausdorff distance for each completion method. In this metric, the CNN completions are significantly closer to the ground truth than are the partial and the mirrored completions. 

\begin{figure}[t]
\centering
	\begin{subfigure}[t]{1.1in}
		\centering
		\includegraphics[width=1.1in]{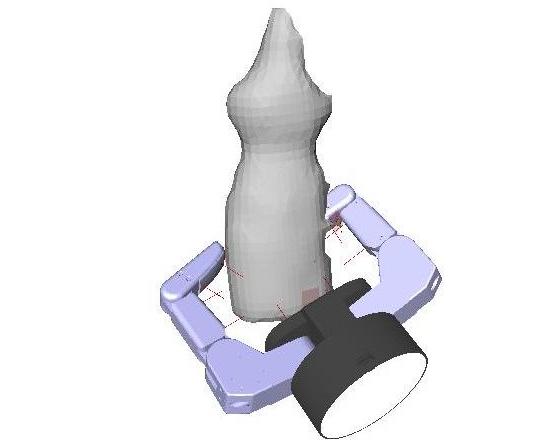}
		\label{fig:p0}	
		\caption{Partial planned}
	\end{subfigure}
	\begin{subfigure}[t]{1.1in}
		\centering
		\includegraphics[width=1.1in]{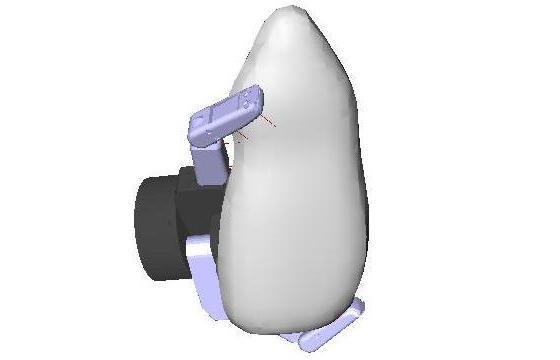}
		\label{fig:p1}
		\caption{Mirrored planned}
	\end{subfigure}
	\begin{subfigure}[t]{1.1in}
		\centering
		\includegraphics[width=1.1in]{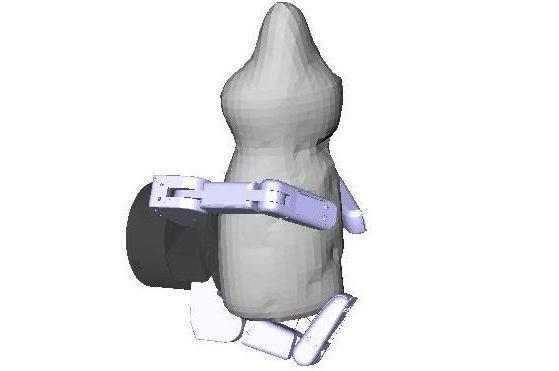}
		\label{fig:p2}	
		\caption{Ours planned}
	\end{subfigure}
	\begin{subfigure}[t]{1.1in}
		\centering
		\includegraphics[width=1.1in]{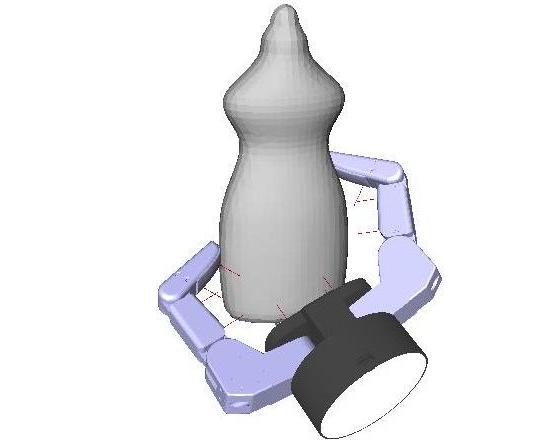}
		\caption{Partial executed}\label{fig:e0}	
	\end{subfigure}
	\begin{subfigure}[t]{1.1in}
		\centering
		\includegraphics[width=1.1in]{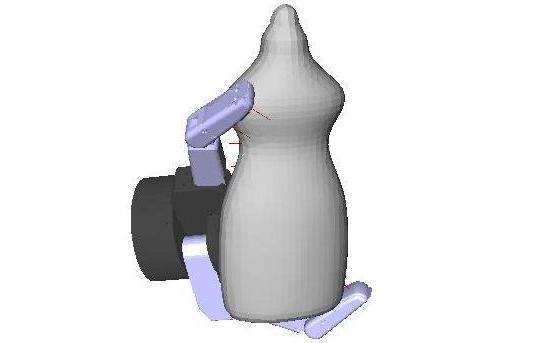}
		\caption{Mirrored executed}\label{fig:e1}
	\end{subfigure}
		\begin{subfigure}[t]{1.1in}
		\centering
		\includegraphics[width=1.1in]{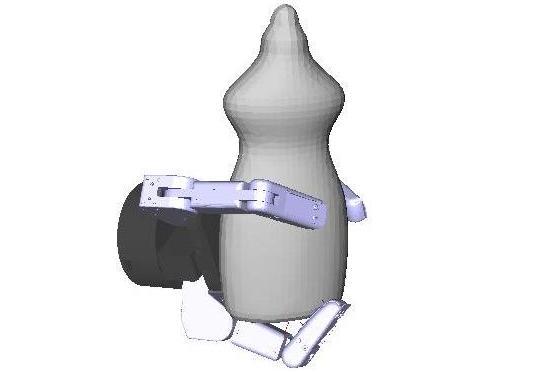}
		\caption{Ours executed}\label{fig:e2}
	\end{subfigure}

\caption{Top Row: Planned Grasps using variety of completions methods. Bottom Row: Grasps from the top row executed on the Ground Truth object. Notice both the partial and mirrored completions' planned and executed grasps differ whereas our method shows fidelity between the planned and executed grasps.} 
\label{fig:sim_example} 
\end{figure} 


The completions are also compared using a measure of geodesic divergence\cite{hamza2003geodesic}. A geodesic shape descriptor is computed for each mesh. A probability density function is then computed for each mesh by considering the shape descriptor as a random distribution and approximating the distribution using a Gaussian mixture model. The probability density functions for each completion are compared with that of the ground truth mesh using the Jenson-Shannon divergence. Table \ref{tab:Geodesic} shows the mean of the divergences for each completion method. Here, our method outperforms all other completion methods.

Across all metrics, our method results in more accurate completions than the other general completion approaches. 

\subsection{Comparison to Database Driven Methods}
\label{sec:Database_comparison_results}



In addition, we evaluated a RANSAC-based approach\cite{papazov2010efficient} on the Training Views of the YCB dataset using the same metrics. This corresponds to a highly constrained environment containing only a very small number of objects which are known ahead of time. It is not possible to load 484 objects into the RANSAC framework, so a direct comparison to our method involving the large number of objects we train on is not possible. In fact, the inability of RANSAC-based methods to scale to large databases of objects is one of the motivations of our work. However, we compared our method to a very small RANSAC using only 14 objects, and our method performs comparably to the RANSAC approach even on objects in its database, while having the additional abilities to train on far more objects and generalize to novel objects: Jaccard (Ours: 0.771, RANSAC: \textbf{0.8566}), Hausdorff (Ours: 3.6, RANSAC: \textbf{3.1}), geodesic (Ours: \textbf{0.0867}, RANSAC: 0.1245). Our approach significantly outperforms the RANSAC approach when encountering an object that neither method has seen before (Holdout Models): Jaccard (Ours: \textbf{0.6496}, RANSAC: 0.4063), Hausdorff (Ours: \textbf{5.9}, RANSAC: 20.4), geodesic (Ours: \textbf{0.1412}, RANSAC: 0.4305). The RANSAC based approach's performance on the Holdout Models is also worse than that of the mirrored or partial completion methods on both the geodesic and Hausdorff metrics.

\subsection{Simulation Based Grasp Comparison}

In order to evaluate our framework's ability to enable grasp planning, the system was tested in simulation using the same completions from Sec \ref{sec:Completion_results}, allowing us to quickly plan and evaluate over 24,000 grasps. GraspIt! was used to plan grasps on all of the completions of the objects by uniformly sampling different approach directions. These grasps were then executed, not on the completed object, but on the ground truth meshes in GraspIt!. In order to simulate a real-world grasp execution, the completion was removed from GraspIt! and the ground truth object was inserted in its place. Then the hand was placed 20cm backed off from the ground truth object along the approach direction of the grasp. The spread angle of the fingers was set, and the hand was moved along the approach direction of the planned grasp either until contact was made or the grasp pose was reached. At this point, the fingers closed to the planned joint values. Then each finger continued to close until either contact was made with the object or the joint limits were reached. Fig. \ref{fig:sim_example} shows several grasps and their realized executions for different completion methods. Visualizations of the simulation results for the entire YCB and Grasp Datasets are available at {\bf http://shapecompletiongrasping.cs.columbia.edu}

Table \ref{tab:sim_grasp_results} shows the differences between the planned and realized joint states as well as the difference in pose of the base of the end effector between the planned and realized grasps. Using our method caused the end effector to end up closer to its intended location in terms of both joint space and the palm's cartesian position.

\subsection{Performance on Real Hardware}

\begin{table}
\vspace{2mm}
\centering
\begin{tabular}{|c|c|c|c|c|c|}
\hline
\multirow{2}{*}{\textbf{View}} & \multirow{2}{*}{\textbf{Error}} & \multicolumn{4}{c|}{\textbf{Completion Type}}\tabularnewline
\cline{3-6}
 &  & \textbf{Partial} & \textbf{Mirror} & \textbf{Ours} & \textbf{RANSAC}\tabularnewline
\hline
\hline
\textbf{Training}  & \textbf{Joint ($^{\circ}$)} & 6.09$^{\circ}$ & 4.20$^{\circ}$ & \textbf{1.75$^{\circ}$} & 1.83$^{\circ}$ \tabularnewline
\cline{2-6}
\textbf{View}  & \textbf{Pose (mm)} & 16.0 & 11.5 & \textbf{4.3}& 7.3\tabularnewline
\hline
\hline
\textbf{Holdout} & \textbf{Joint ($^{\circ}$)} & 6.27$^{\circ}$ & 4.05$^{\circ}$ & 1.80$^{\circ}$& \textbf{1.69$^{\circ}$} \tabularnewline
\cline{2-6}
\textbf{View} & \textbf{Pose (mm)} & 20.8 & 15.6 & \textbf{6.7} & 7.4\tabularnewline
\hline
\hline
\textbf{Holdout} & \textbf{Joint ($^{\circ}$)} & 7.59$^{\circ}$ & 5.82$^{\circ}$ & \textbf{4.56$^{\circ}$}&6.86$^{\circ}$ \tabularnewline
\cline{2-6}
\textbf{Model} & \textbf{Pose (mm)} & 18.3 & 15.0 & \textbf{13.2}& 29.25 \tabularnewline
\hline
\end{tabular}

\caption{Results from simulated grasping experiments. Joint Err. is the mean difference between planned and realized grasps per joint in degrees. Pose Err. is the mean difference between planned and realized grasp pose in millimeters. For both metrics smaller is better. }
\label{tab:sim_grasp_results} 
\end{table}

\begin{table}
\centering
\begin{tabular}{|c|c|c|c|c| }
\hline
\multicolumn{1}{|c|}{\begin{tabular}[c]{@{}c@{}}\textbf{Completion} \\  \textbf{Method}\end{tabular}} 
& \multicolumn{1}{c|}{\begin{tabular}[c]{@{}c@{}}\textbf{Grasp Success} \\  \textbf{Rate (\%)}\end{tabular}} 
& \multicolumn{1}{c|}{\begin{tabular}[c]{@{}c@{}}\textbf{Joint Error} \\  \textbf{(degrees)}\end{tabular}} 
& \multicolumn{1}{c|}{\begin{tabular}[c]{@{}c@{}}\textbf{Completion} \\  \textbf{Time (s)}\end{tabular}} 
 \\ \hline

	Partial & 71.43& 9.156$^{\circ}$& \textbf{0.545} \\ \hline
	Mirror & 73.33& 8.067$^{\circ}$& 1.883 \\ \hline
	Ours & \textbf{93.33}& \textbf{7.276}$^{\circ}$& 2.426 \\ \hline
\end{tabular}

\caption{Grasp Success Rate shows the percentage of succesful grasp attempts. Joint Error shows the mean difference in degrees between the planned and executed grasp joint values. Completion Time shows how long specified metric took to create a mesh of the partial view.}
\label{tab:live_grasp_results} 
\end{table}

\begin{figure}[t]
\vspace{2mm}
\centering
\includegraphics  [width=.45\textwidth] {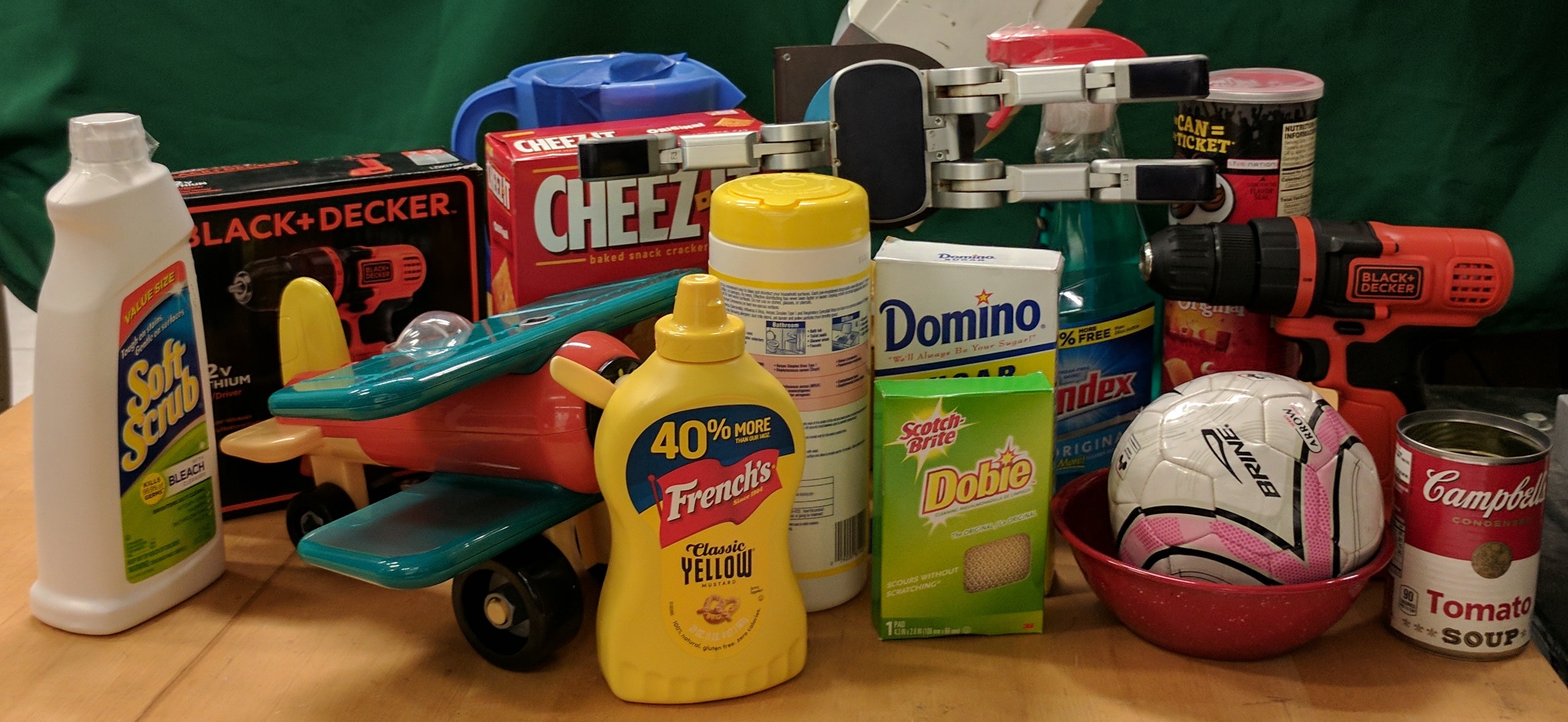}
\caption{Barrett Hand(BH8-280), StaubliTX60 Arm, and experiment objects.}
\label{fig:experimental_setup}
\end{figure}

In order to further evaluate our framework, the system was used in an end-to-end manner using actual robotic hardware to execute grasps planned via the different completion methods described above. The 15 objects used are shown in Fig. \ref{fig:experimental_setup}. 
For each object, we ran the arm once using each completion method. The results are shown in Table \ref{tab:live_grasp_results}. Our method enabled a 20\% improvement over the general shape completion methods in terms of grasp success rate, and resulted in executed grasps closer to the planned grasps as shown by the lower joint error. 

\begin{figure}[t]
\centering
\vspace{2mm}
	\begin{subfigure}[t]{1.6in}
		\centering
		\includegraphics[width=1.6in, height=1.1in]{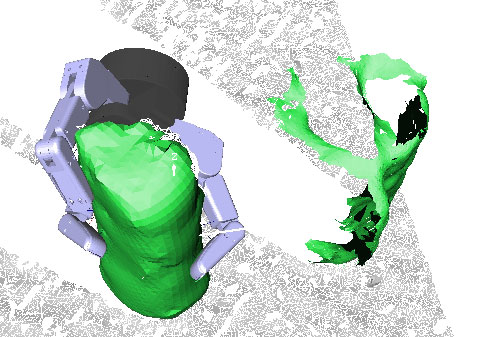}
		\caption{Planned Grasp}\label{fig:pringles_grasp_planned_unfilled}
	\end{subfigure}
	\begin{subfigure}[t]{1.6in}
		\centering
		\includegraphics[width=1.6in, height=1.1in]{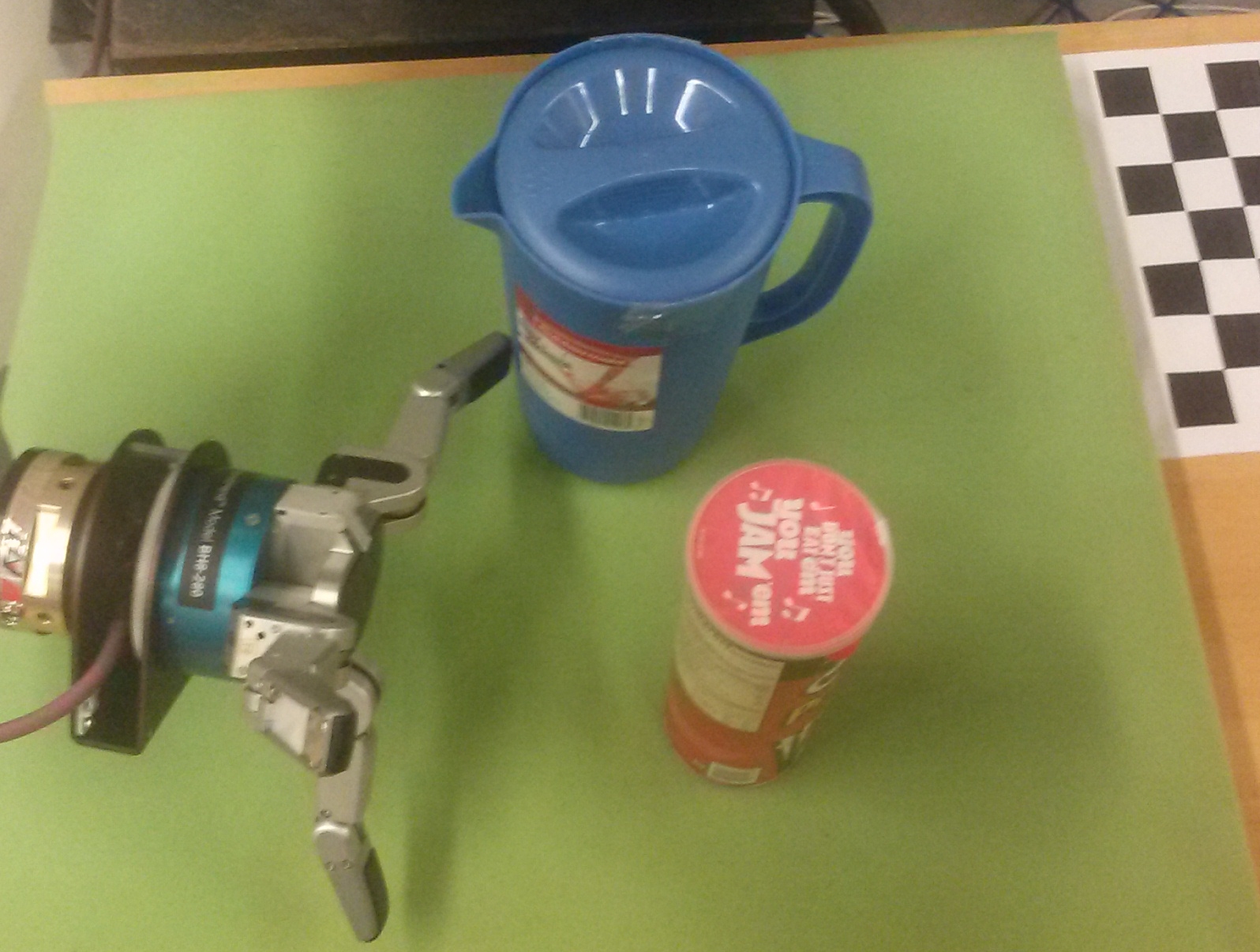}
		\caption{Accidental Collision}\label{fig:pringles_grasp_live}
	\end{subfigure}
	\begin{subfigure}[t]{1.6in}
		\centering
		\includegraphics[width=1.6in, height=1.1in]{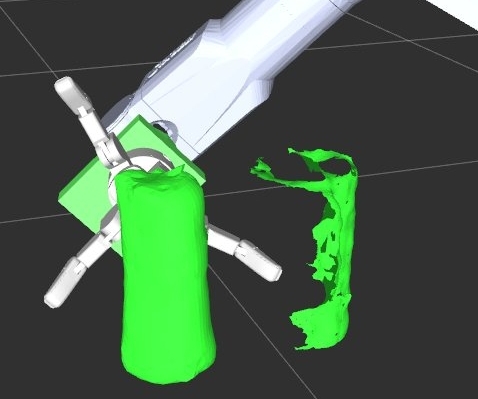}
		\caption{Unfilled Planning Scene}\label{fig:pringles_collision_check}
	\end{subfigure}
	\begin{subfigure}[t]{1.6in}
		\centering
		\includegraphics[width=1.6in, height=1.1in]{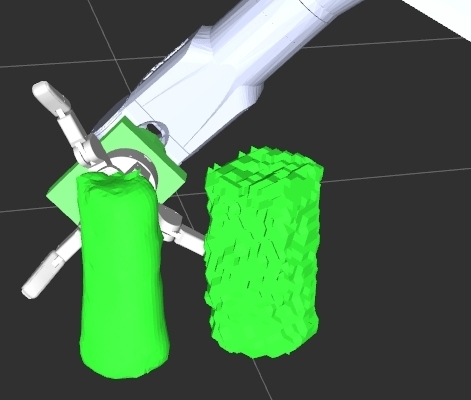}
		\caption{Filled Planning Scene}\label{fig:pringles_filled_fast}
	\end{subfigure}
\caption{The system can be used to quickly complete obstacles that are to be avoided. The arm fails to execute the planned grasp (a), resulting in collision shown in (b). The collision with the non-target object occurred due to a poor planning scene as shown in (c), but the CNN without post-processing can be used to fill the planning scene allowing the configuration to be correctly marked as invalid as shown in (d).} 
\label{fig:hand_collision} 
\end{figure}

\subsection{Crowded Scene Completion}

A scene often contains objects that are not to be manipulated and only require completion in order to be successfully avoided. In this case, the output of our CNN can be run directly through marching cubes without post-processing to quickly create a mesh of the object. Fig. \ref{fig:hand_collision}(a) shows a grasp planned using only the partial mesh for the object near the grasp target. Figs. \ref{fig:hand_collision}(b) and \ref{fig:hand_collision}(c) show the robotic hand crashing into one of the nearby objects when attempting to execute the grasp. The failure is caused by an incomplete planning scene. Fig. \ref{fig:hand_collision}(d) shows the scene with the nearby objects completed, though without smoothing. With this fuller picture, the planner accurately flags this grasp as unreachable. The time requirement for the scene completion is:
$$T_{completion} = T_{segment} + T_{target} + T_{non\_target}*n$$

\noindent
with Segmentation Time ($T_{segment}$), Target Completion Time ($T_{target}$), Non Target Completion Time($T_{non\_target}$) and Number of Non Target Objects ($n$). Our system has the ability to quickly fill in occluded regions of the scene, and selectively spend more time generating detailed completions on specific objects to be manipulated. Average completion times in seconds from 15 runs are $T_{segment} = 0.119$, $T_{target} = 2.136$, and $T_{non\_target} = 0.142$.




\section{Conclusion and Future Work}

This work presents a framework to train and utilize a CNN to complete and mesh an object observed from a single point of view, and then plan grasps on the completed object. The completion system is fast, with completions available in a matter of milliseconds, and post processed completions suitable for grasp planning available in several seconds. The dataset and code are open source and available for other researchers to use. It has also been demonstrated that our completions are better than more naive approaches in terms of a variety of metrics including those specific to grasp planning. In addition, grasps planned on completions generated using our method are more often successful and result in executed grasps closer to the intended hand configuration than grasps planned on completions from the other methods.  

Several future avenues of research include: the use of Generative Adversarial Networks (GANs)\cite{goodfellow2014generative} for training, migrating to a larger object dataset such as ShapeNet\cite{chang2015shapenet}, and using the completed object as a query to retrieve grasps planned on similar objects as done with the Columbia Grasp Database\cite{goldfeder2009columbia} and DexNet\cite{mahler2016dex}.


\bibliographystyle{IEEEtran}
\bibliography{references}

\begin{thebibliography}{10}
\providecommand{\url}[1]{#1}
\csname url@samestyle\endcsname
\providecommand{\newblock}{\relax}
\providecommand{\bibinfo}[2]{#2}
\providecommand{\BIBentrySTDinterwordspacing}{\spaceskip=0pt\relax}
\providecommand{\BIBentryALTinterwordstretchfactor}{4}
\providecommand{\BIBentryALTinterwordspacing}{\spaceskip=\fontdimen2\font plus
\BIBentryALTinterwordstretchfactor\fontdimen3\font minus
  \fontdimen4\font\relax}
\providecommand{\BIBforeignlanguage}[2]{{%
\expandafter\ifx\csname l@#1\endcsname\relax
\typeout{** WARNING: IEEEtran.bst: No hyphenation pattern has been}%
\typeout{** loaded for the language `#1'. Using the pattern for}%
\typeout{** the default language instead.}%
\else
\language=\csname l@#1\endcsname
\fi
#2}}
\providecommand{\BIBdecl}{\relax}
\BIBdecl

\bibitem{lecun}
Y.~LeCun and Y.~Bengio, ``Convolutional networks for images, speech, and time
  series,'' in \emph{Handbk. of brain theory \& neural networks}, 1995.

\bibitem{lorensen1987marching}
W.~E. Lorensen and H.~E. Cline, ``Marching cubes: A high resolution 3d surface
  construction algorithm,'' in \emph{ACM siggraph computer graphics}, vol.~21,
  no.~4.\hskip 1em plus 0.5em minus 0.4em\relax ACM, 1987, pp. 163--169.

\bibitem{chang2015shapenet}
A.~X. Chang, T.~Funkhouser, L.~Guibas, P.~Hanrahan, Q.~Huang, Z.~Li,
  S.~Savarese, M.~Savva, S.~Song, H.~Su \emph{et~al.}, ``Shapenet: An
  information-rich 3d model repository,'' \emph{arXiv preprint
  arXiv:1512.03012}, 2015.

\bibitem{bohg2011mind}
J.~Bohg, M.~Johnson-Roberson, B.~Le{\'o}n, J.~Felip, X.~Gratal,
  N.~Bergstr{\"o}m, D.~Kragic, and A.~Morales, ``Mind the gap-robotic grasping
  under incomplete observation,'' in \emph{ICRA}.\hskip 1em plus 0.5em minus
  0.4em\relax IEEE, 2011, pp. 686--693.

\bibitem{quispe2015exploiting}
A.~H. Quispe, B.~Milville, M.~A. Guti{\'e}rrez, C.~Erdogan, M.~Stilman,
  H.~Christensen, and H.~B. Amor, ``Exploiting symmetries and extrusions for
  grasping household objects,'' in \emph{ICRA}, 2015, pp. 3702--3708.

\bibitem{schiebener2016heuristic}
D.~Schiebener, A.~Schmidt, N.~Vahrenkamp, and T.~Asfour, ``Heuristic 3d object
  shape completion based on symmetry and scene context,'' in \emph{IROS}.\hskip
  1em plus 0.5em minus 0.4em\relax IEEE, 2016, pp. 74--81.

\bibitem{rennie2016dataset}
C.~Rennie, R.~Shome, K.~E. Bekris, and A.~F. De~Souza, ``A dataset for improved
  rgbd-based object detection and pose estimation for warehouse
  pick-and-place,'' \emph{IEEE Robotics and Automation Letters}, vol.~1, no.~2,
  pp. 1179--1185, 2016.

\bibitem{papazov2010efficient}
C.~Papazov and D.~Burschka, ``An efficient ransac for 3d object recognition in
  noisy and occluded scenes,'' in \emph{Asian Conference on Computer
  Vision}.\hskip 1em plus 0.5em minus 0.4em\relax Springer, 2010, pp. 135--148.

\bibitem{hinterstoisser2011multimodal}
S.~Hinterstoisser, S.~Holzer, C.~Cagniart, S.~Ilic, K.~Konolige, N.~Navab, and
  V.~Lepetit, ``Multimodal templates for real-time detection of texture-less
  objects in heavily cluttered scenes,'' in \emph{ICCV), 2011}.\hskip 1em plus
  0.5em minus 0.4em\relax IEEE, 2011, pp. 858--865.

\bibitem{wu20143d}
Z.~Wu, S.~Song, A.~Khosla, X.~Tang, and J.~Xiao, ``3{D} shapenets for 2.5{D}
  object recognition and next-best-view prediction,'' \emph{arXiv preprint
  arXiv:1406.5670}, 2014.

\bibitem{wu20153d}
Z.~Wu, S.~Song, A.~Khosla, F.~Yu, L.~Zhang, X.~Tang, and J.~Xiao, ``3d
  shapenets: A deep representation for volumetric shapes,'' in \emph{CVPR},
  2015, pp. 1912--1920.

\bibitem{firman2016structured}
M.~Firman, O.~Mac~Aodha, S.~Julier, and G.~J. Brostow, ``Structured prediction
  of unobserved voxels from a single depth image,'' in \emph{CVPR}, 2016, pp.
  5431--5440.

\bibitem{rock2015completing}
J.~Rock, T.~Gupta, J.~Thorsen, J.~Gwak, D.~Shin, and D.~Hoiem, ``Completing 3d
  object shape from one depth image,'' in \emph{CVPR}, 2015, pp. 2484--2493.

\bibitem{tulsiani2016learning}
S.~Tulsiani, A.~Kar, J.~Carreira, and J.~Malik, ``Learning category-specific
  deformable 3d models for object reconstruction,'' \emph{IEEE transactions on
  pattern analysis and machine intelligence}, 2016.

\bibitem{choy20163d}
C.~B. Choy, D.~Xu, J.~Gwak, K.~Chen, and S.~Savarese, ``3d-r2n2: A unified
  approach for single and multi-view 3d object reconstruction,'' in
  \emph{ECCV}.\hskip 1em plus 0.5em minus 0.4em\relax Springer, 2016, pp.
  628--644.

\bibitem{li2015Database}
Y.~Li, A.~Dai, L.~Guibas, and M.~Nie{\ss}ner, ``Database-assisted object
  retrieval for real-time 3d reconstruction,'' in \emph{Computer Graphics
  Forum}, vol.~34, no.~2.\hskip 1em plus 0.5em minus 0.4em\relax Wiley Online
  Library, 2015, pp. 435--446.

\bibitem{mahler2016dex}
J.~Mahler, F.~T. Pokorny, B.~Hou, M.~Roderick, M.~Laskey, M.~Aubry,
  K.~Kohlhoff, T.~Kr{\"o}ger, J.~Kuffner, and K.~Goldberg, ``Dex-net 1.0: A
  cloud-based network of 3d objects for robust grasp planning using a
  multi-armed bandit model with correlated rewards,'' in \emph{ICRA,
  2016}.\hskip 1em plus 0.5em minus 0.4em\relax IEEE, 2016, pp. 1957--1964.

\bibitem{goldfeder2009columbia}
C.~Goldfeder, M.~Ciocarlie, H.~Dang, and P.~K. Allen, ``The columbia grasp
  database,'' in \emph{ICRA}.\hskip 1em plus 0.5em minus 0.4em\relax IEEE,
  2009, pp. 1710--1716.

\bibitem{varley2015generating}
J.~Varley, J.~Weisz, J.~Weiss, and P.~Allen, ``Generating multi-fingered
  robotic grasps via deep learning,'' in \emph{IROS}, 2015, pp. 4415--4420.

\bibitem{xiang_wacv14}
Y.~Xiang, R.~Mottaghi, and S.~Savarese, ``Beyond pascal: A benchmark for 3d
  object detection in the wild,'' in \emph{WACV}, 2014.

\bibitem{calli2015ycb}
B.~Calli, A.~Singh, A.~Walsman, S.~Srinivasa, P.~Abbeel, and A.~M. Dollar,
  ``The ycb object and model set: Towards common benchmarks for manipulation
  research,'' in \emph{Advanced Robotics (ICAR), 2015 International Conference
  on}.\hskip 1em plus 0.5em minus 0.4em\relax IEEE, 2015, pp. 510--517.

\bibitem{kappler2015leveraging}
D.~Kappler, J.~Bohg, and S.~Schaal, ``Leveraging big data for grasp planning,''
  in \emph{ICRA}.\hskip 1em plus 0.5em minus 0.4em\relax IEEE, 2015, pp.
  4304--4311.

\bibitem{min2004binvox}
P.~Min, ``Binvox, a 3{d} mesh voxelizer,'' 2004.

\bibitem{koenig2004design}
N.~Koenig and A.~Howard, ``Design and use paradigms for {Gazebo}, an
  open-source multi-robot simulator,'' in \emph{IROS}, vol.~3.\hskip 1em plus
  0.5em minus 0.4em\relax IEEE, 2004, pp. 2149--2154.

\bibitem{Keras2015}
F.~Chollet, ``Keras,'' \url{https://github.com/fchollet/keras}, 2015.

\bibitem{bergstra2010theano}
J.~Bergstra, O.~Breuleux, F.~Bastien, P.~Lamblin, R.~Pascanu, G.~Desjardins,
  J.~Turian, D.~Warde-Farley, and Y.~Bengio, ``Theano: a cpu and gpu math
  expression compiler,'' in \emph{Proceedings of the Python for scientific
  computing conference (SciPy)}, vol.~4.\hskip 1em plus 0.5em minus 0.4em\relax
  Austin, TX, 2010.

\bibitem{bastien2012theano}
F.~Bastien, P.~Lamblin, R.~Pascanu, J.~Bergstra, I., A.~Bergeron, N.~Bouchard,
  D.~Warde-Farley, and Y.~Bengio, ``Theano: new features and speed
  improvements,'' \emph{arXiv:1211.5590}, 2012.

\bibitem{kingma2014}
D.~Kingma and J.~Ba, ``Adam: A method for stochastic optimization,''
  \emph{arXiv preprint arXiv:1412.6980}, 2014.

\bibitem{he2015}
K.~He, X.~Zhang, S.~Ren, and J.~Sun, ``Delving deep into rectifiers: Surpassing
  human-level performance on imagenet classification,'' \emph{arXiv preprint
  arXiv:1502.01852}, 2015.

\bibitem{glorot2010}
X.~Glorot and Y.~Bengio, ``Understanding the difficulty of training deep
  feedforward neural networks,'' in \emph{Proceedings of the 13th AISTATS},
  2010, pp. 249--256.

\bibitem{Rusu_ICRA2011_PCL}
R.~B. Rusu and S.~Cousins, ``{3D is here: Point Cloud Library (PCL)},'' in
  \emph{{ICRA}}, Shanghai, China, May 9-13 2011.

\bibitem{lempitsky2010surface}
V.~Lempitsky, ``Surface extraction from binary volumes with higher-order
  smoothness,'' in \emph{Computer Vision and Pattern Recognition (CVPR), 2010
  IEEE Conference on}.\hskip 1em plus 0.5em minus 0.4em\relax IEEE, 2010, pp.
  1197--1204.

\bibitem{miller2004graspit}
A.~T. Miller and P.~K. Allen, ``Graspit! a versatile simulator for robotic
  grasping,'' \emph{IEEE R\&A Magazine}, vol.~11, no.~4, pp. 110--122, 2004.

\bibitem{sucan2013moveit}
I.~A. Sucan and S.~Chitta, ``Moveit!'' \emph{http://moveit.ros.org}, 2013.

\bibitem{cignoni2008meshlab}
P.~Cignoni, M.~Corsini, and G.~Ranzuglia, ``Meshlab: an open-source 3d mesh
  processing system,'' \emph{Ercim news}, vol.~73, pp. 45--46, 2008.

\bibitem{hamza2003geodesic}
A.~B. Hamza and H.~Krim, ``Geodesic object representation and recognition,'' in
  \emph{International conference on discrete geometry for computer
  imagery}.\hskip 1em plus 0.5em minus 0.4em\relax Springer, 2003, pp.
  378--387.

\bibitem{goodfellow2014generative}
I.~Goodfellow, J.~Pouget-Abadie, M.~Mirza, B.~Xu, D.~Warde-Farley, S.~Ozair,
  A.~Courville, and Y.~Bengio, ``Generative adversarial nets,'' in \emph{Adv.
  in neural information processing systems}, 2014, pp. 2672--2680.

\end{thebibliography}

\end{document}